\documentclass[conference]{IEEEtran}

\usepackage{cite}
\usepackage{graphicx}
\usepackage{amsmath}
\usepackage{amssymb}
\usepackage{xfrac}
\usepackage{algorithmic}
\usepackage{array}
\usepackage{url}
\usepackage{comment}
\usepackage{xcolor}

\makeatletter
\def\blfootnote{\xdef\@thefnmark{}\@footnotetext}
\makeatother

\begin{document}

\title{Morse Code Datasets for Machine Learning}

\author{\IEEEauthorblockN{Sourya Dey, Keith M.~Chugg and Peter A.~Beerel}
\IEEEauthorblockA{Ming Hsieh Department of Electrical Engineering\\
University of Southern California\\
Los Angeles, California 90089, USA\\
\{souryade, chugg, pabeerel\}@usc.edu}}

\maketitle


\begin{abstract}
We present an algorithm to generate synthetic datasets of tunable difficulty on classification of Morse code symbols for supervised machine learning problems, in particular, neural networks. The datasets are spatially one-dimensional and have a small number of input features, leading to high density of input information content. This makes them particularly challenging when implementing network complexity reduction methods. We explore how network performance is affected by deliberately adding various forms of noise and expanding the feature set and dataset size. Finally, we establish several metrics to indicate the difficulty of a dataset, and evaluate their merits. The algorithm and datasets are open-source.
\end{abstract}

\blfootnote{\copyright 2018 IEEE\qquad Original IEEE Publication Citation:\\S. Dey, K. M. Chugg and P. A. Beerel, "Morse Code Datasets for Machine Learning," 2018 9th International Conference on Computing, Communication and Networking Technologies (ICCCNT), 2018, pp. 1-7.
doi: 10.1109/ICCCNT.2018.8494011}

\section{Introduction and Prior Work}\label{intro}
Neural networks in machine learning systems are commonly employed to tackle classification problems involving characters or images. In such problems, the neural network (NN) processes an input sample and predicts which \emph{class} it belongs to. The inputs and their classes are drawn from a \emph{dataset}, such as the MNIST \cite{LeCun1998_MNIST} dataset containing images of handwritten digits, or the CIFAR \cite{Krizhevsky2009_CIFAR} and ImageNet \cite{Russakovsky2015_ImageNet} datasets containing images of common objects such as birds and houses. A NN is first trained using numerous examples where the input sample and its class label are both available, then used for inference (i.e. prediction) of the classes of input samples whose class labels are not available. The training stage is data-hungry and typically requires thousands of labeled examples. It is therefore often a challenge to obtain adequate amounts of high quality and accurate data required to sufficiently train a NN. A possible solution is to obtain data by synthetic instead of natural means. Synthetic data are generated using computer algorithms instead of being collected from real-world scenarios. The advantages are that a) computer algorithms can be tuned to mimic real-world settings to desired levels of accuracy, and b) a theoretically unlimited amount of data can be generated by running the algorithm long enough. The effects of dataset size on network performance has been explored in \cite{Weiss2003}, in particular, more inputs are beneficial in reducing overfitting and improving robustness and generalization capabilities of NNs \cite{Goodfellow2016_DLbook,Goodfellow2015_Adversarial}. Synthetic data has been successfully used in problems such as 3D imaging \cite{Peng2015}, point tracking \cite{DeTone2017}, breaking Captchas on popular websites \cite{Le2017}, and augmenting real world datasets \cite{Patki2016}.

This present work introduces a family of synthetic datasets on classifying Morse codewords. Morse code is a system of communication where each letter, number or symbol in a language is represented using a sequence of dots and dashes, separated by spaces. It is widely used to communicate in situations where voice is not possible, such as helping people with disabilities talk \cite{Bakde2012,Yang2003,Yang2001,Luo1996}, or where message transmission needs to be achieved using only 2 states \cite{Ravikumar2016}, or in rehabilitation and education \cite{King1999}. Morse code is a useful skill to learn and there exist cellphone apps designed to train people in its usage \cite{androidmorse,iOSmorse}.

Our work uses feed-forward multi-layer perceptron neural networks to directly classify Morse codewords into 64 character classes comprising letters, numbers and symbols. This is different from previous works such as \cite{Yang2001,Luo2001,Ravikumar2016,Luo1996} which only had 2 classes corresponding to dots and dashes. In particular, \cite{Ravikumar2016} used fuzzy logic on inputs from a microcontroller used in security systems, while \cite{Luo1996} used least mean squares approximation, both to classify dots and dashes. There has also been previous work using time series and recurrent networks to decode English words in Morse code \cite{Hill1992,Aly2000}, while \cite{Li2017} used radial basis function networks to classify characters with 84\% \emph{accuracy}. Accuracy is a common metric for describing the performance of a classification NN and is measured as the percentage of class labels correctly predicted by the NN during inference.

The key contributions of the present work are as follows:
\begin{itemize}
    \item An algorithm (described in Section \ref{algo}) to generate machine learning datasets of varying difficulty. To the best of our knowledge, we are the first to develop an algorithm which can scale the difficulty of machine learning datasets. The difficulty of a dataset can be observed from the accuracy of a NN training on it -- harder datasets lead to lower accuracy, and vice-versa. We discuss techniques to make datasets harder and show corresponding accuracy results in Section \ref{anares}. Encountering harder datasets leads to aggressive exploration of network hyperparameters and learning algorithms, which ultimately increases the robustness of NNs training on them. The algorithm and datasets are open source and available on Github \cite{morse-dataset-github}.
    \item In Section \ref{metrics}, we introduce metrics to quantify the difficulty of a dataset. While some of these arise from information theory, we also come up with a new metric which achieves a high correlation coefficient with the accuracy achieved by NNs on a dataset. Our metrics are a useful way to characterize how hard a dataset is \emph{without} having a NN train on them.
    \item This work is one of few to introduce a spatially 1-dimensional dataset. This is in contrast to the wide array of image and character recognition datasets which are usually 2-dimensional such as MNIST, where each image has width and height, or 3-dimensional such as CIFAR and ImageNet, where each image has width, height and a number of features. The number of spatial dimensions in the input data is important when dealing with low-complexity \emph{sparse} NNs. Previous works \cite{Dey2017_ICANN,Dey2017_Asilomar,Dey2018_ITA,Han2016DC,Han2015,Chen2015,Zhou2016} have focused on making NNs sparse, while keeping the resulting accuracy reduction to a minimum. The family of Morse code datasets described in the present work was designed to test the limits of sparse NNs, as described in Section \ref{sparseresults}. 
\end{itemize}

\section{Generating Algorithm}\label{algo}
We picked 64 class labels for our dataset -- the 26 English letters, the 10 Arabic numerals, and 28 other symbols such as \textbf{(}, \textbf{+}, \textbf{:}, etc. Each of these is represented by a sequence of dots and dashes in Morse code, for example, \textbf{+} is represented as \mbox{$\bullet$\textbf{ --- }$\bullet$\textbf{ --- }$\bullet$}. So as to mimic a real-world scenario in our algorithm, we imagined a human or a Morse code machine writing out this sequence within a frame of fixed size. Wherever the pen or electronic instrument touches is darkened and has a high intensity, indicating the presence of dots and dashes, while the other parts are left blank (i.e. spaces).

\subsubsection{Step 1 -- Frame Partitioning}
For our algorithm, each Morse codeword lies in a frame which is a vector of 64 values. Within the frame, the length of a sequence having consecutive similar values is used to differentiate between a dot and a dash. In the baseline dataset, a dot can be 1-3 values wide and a dash 4-9. This is in accordance with international Morse code regulations \cite{morseregulations2009} where the size or duration of a dash is around 3 times that of a dot. The space between a dot and a dash can have a length of 1-3 values. The exact length of a dot, dash or space is chosen from these ranges according to a uniform probability distribution. This is to mimic the human writer who is not expected to make each symbol have a consistent length, but can be expected to make dots and spaces around the same size, and dashes longer than them. The baseline dataset has no leading spaces before the 1st dot or dash, i.e. the codeword starts from the left edge of the frame. There are trailing spaces to fill up the right side of the frame after all the dots and dashes are complete.

\subsubsection{Step 2 -- Assigning Values for Intensity Levels}
All values in the frame are initially real numbers in the range $[0,16]$ and indicate the intensity of that point in the frame. For dots and dashes, the values are drawn from a normal distribution with mean $\mu=12$ and standard deviation $\sigma=\sfrac{4}{3}$. The idea is to have the ‘six-sigma’ range from $(12-3\times\sfrac{4}{3})=8$ to $(12+3\times\sfrac{4}{3})=16$. This ensures that any value making up a dot or a dash will lie in the upper half of possible values, i.e. in the range $[8,16]$. The value of a space is exactly 0. Once again, these conditions mimic the human or machine writer who is not expected to have consistent intensity for every dot and dash, but can be expected to not let the writing instrument touch portions of the frame which are spaces.

\subsubsection{Step 3 -- Noising}
\begin{figure}[!t]
\centering
\includegraphics[width = 1.0\linewidth]{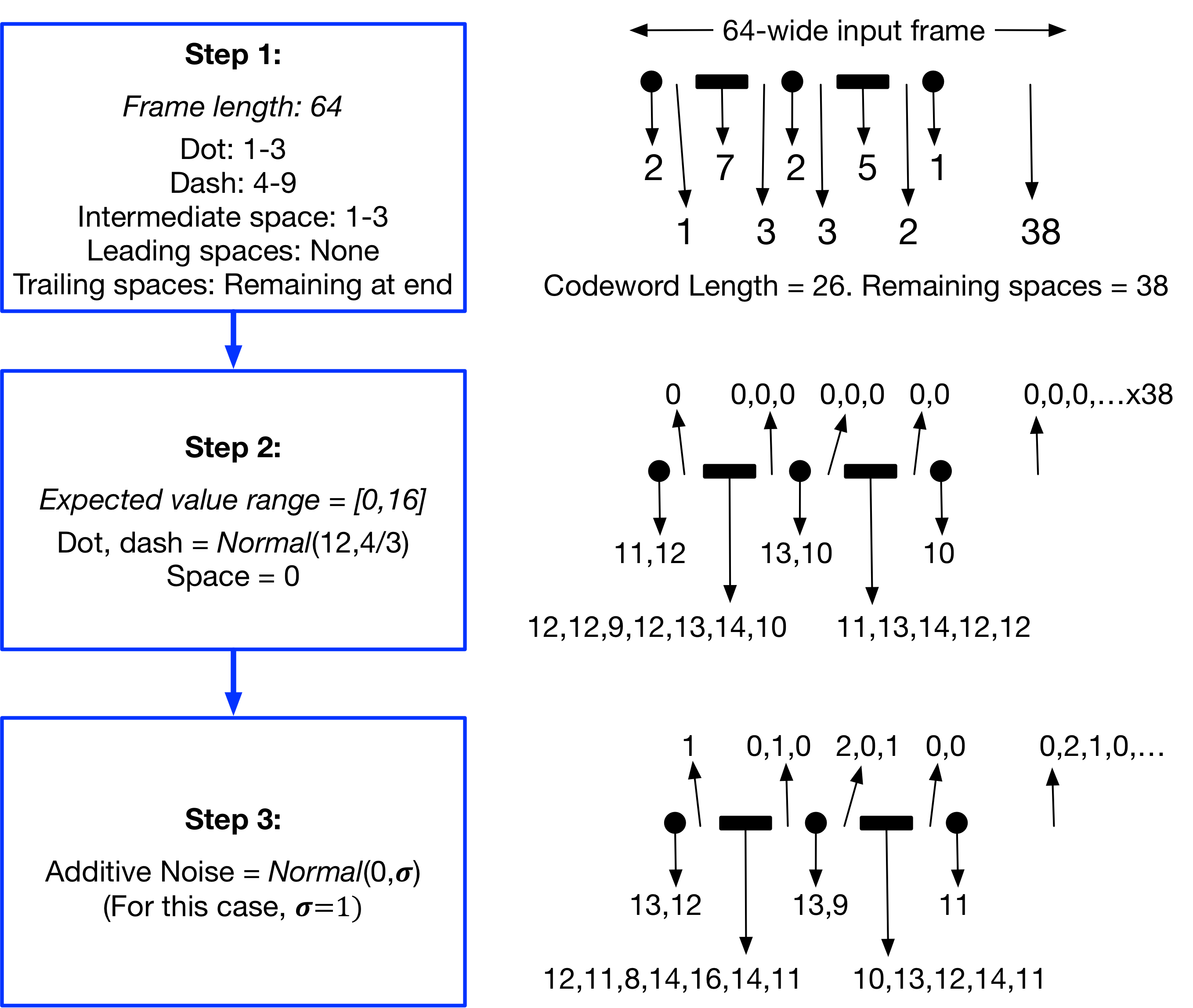}
\caption{Generating the Morse codeword \mbox{$\bullet$\textbf{ --- }$\bullet$\textbf{ --- }$\bullet$} corresponding to the \textbf{+} symbol. The first 3 steps, prior to normalizing, are shown. Only integer values are shown for convenience, however, the values can and generally will be fractional. \emph{Normal}$(\mu,\sigma)$ denotes a normal distribution with mean = $\mu$, standard deviation = $\sigma$. For this figure, $\sigma=1$.}
\label{fig-algo}
\end{figure}

Noise in input samples is often deliberately injected as a means of avoiding overfitting in NNs \cite{Goodfellow2016_DLbook}, and has been shown to be superior to other methods of avoiding overfitting \cite{Zur2009}. This was, however, the secondary reason behind our experimenting with noise. The primary reason was to deliberately make the data hard to classify and test the limits of different NNs processing it. Noise can be thought of as a human accidentally varying the intensity of writing the Morse codeword, or a Morse communication channel having noise. The baseline dataset has no noise, while others have additive noise from a mean-zero normal distribution applied to them. Fig. \ref{fig-algo} shows the 3 steps up to this point. Finally, all the values are normalized to lie within the range $[0,1]$ with precision of 3 decimal places.

\subsubsection{Step 4 -- Mass Generation}
Steps 1-3 describe the generation of 1 input sample corresponding to some particular class label. This can be repeated as many times as required for each of the 64 class labels. This demonstrates a key advantage of synthetic over real-world data -- the ability to generate an arbitrary amount of data having an arbitrary prior probability distribution over its classes. The baseline dataset has 7,000 examples for each class, for a total of 448,000 examples.

\subsection{Variations and Difficulty Scaling}\label{variations}

\begin{figure*}[!t]
\centering
\includegraphics[width = 0.85\linewidth]{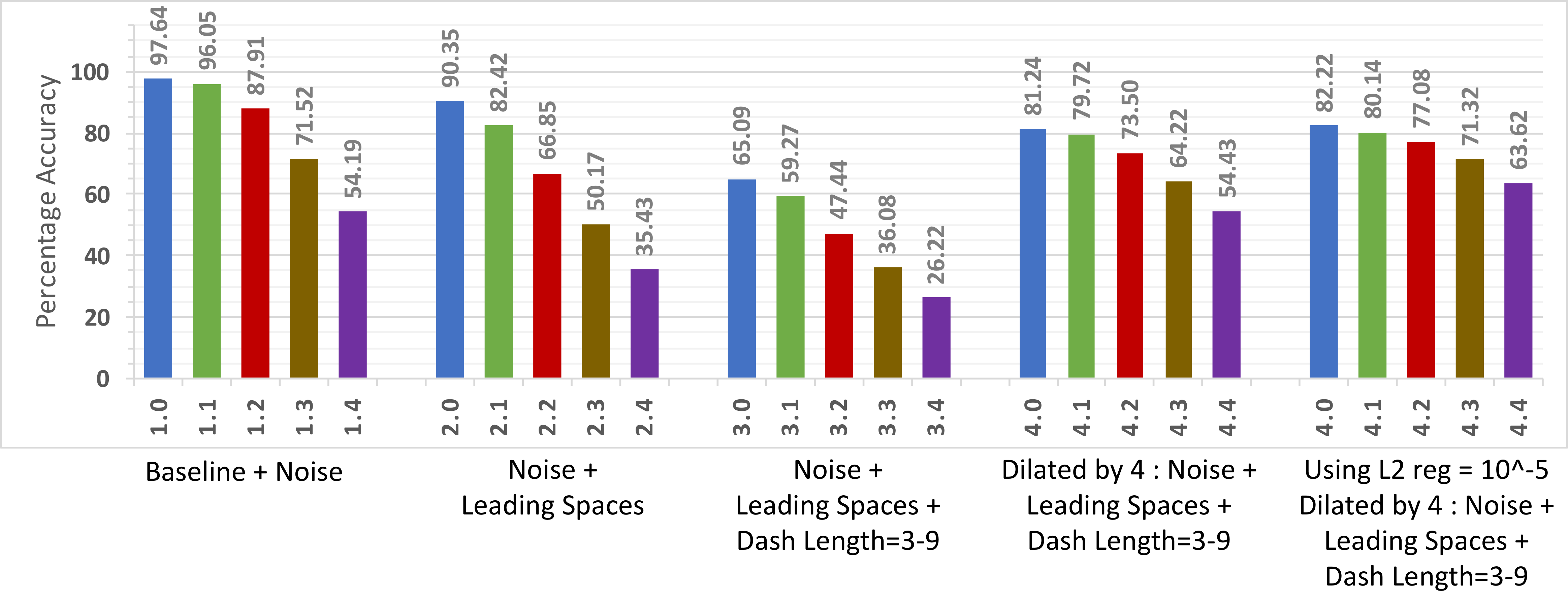}
\caption{Percentage accuracies obtained by the NN described in \ref{netsetup} on the test subset of the datasets described in \ref{variations}. The rightmost set of bars corresponds to \emph{Morse 4.}$\sigma$ with L2 regularization.}
\label{fig-fc_results}
\end{figure*}

The baseline dataset is as described so far, except that $\sigma=0$, i.e. it has no additive noise. We experimented with the following variations in datasets:
\begin{enumerate}
    \item Baseline with \textbf{additive noise} = \emph{Normal}$(0,\sigma)$, $\sigma \in \{0,1,2,3,4\}$. These are called \emph{Morse 1.}$\sigma$, i.e. \emph{1.0} to \emph{1.4}, where \emph{1.0} is the baseline. 
    \item Instead of having the codeword start from the left edge of the frame, we introduced a random number of \textbf{leading spaces}. For example, in Fig. \ref{fig-algo}, the codeword occupies a length of 26 values. The remaining 38 space values can be randomly divided between leading and trailing spaces. This increases the difficulty of the dataset since no particular set of neurons are expected to be learning dots and dashes as the actual codeword could be anywhere in the frame. Just like variation 1, we added noise and call these datasets \emph{Morse 2.}$\sigma$, $\sigma \in \{0,1,2,3,4\}$.
    \item There is no overlap between the lengths of dots and dashes in the datasets described so far. The difficulty can be increased by making \textbf{dash length = 3-9 values}, which is exactly according to the convention of having dash length thrice of dot length. This means that dashes can masquerade as dots and spaces, and vice-versa. This is done on top of introducing leading spaces. These datasets are called \emph{Morse 3.}$\sigma$, $\sigma$ being as before.
    \item The Morse datasets only have 64 inputs, which is quite small compared to others such as MNIST (784 inputs), CIFAR (3072 inputs), or ImageNet (150,528 inputs). This makes the Morse datasets hard to classify since there is less redundancy in inputs, so a given amount of noise will lead to greater reductions in signal-to-noise ratio (SNR) compared to other datasets. To make the Morse datasets easier, we introduced \textbf{dilation} by a factor of 4. This is done by scaling all lengths in variation 3 by a factor of 4, i.e. frame length is 256, dot sizes and space sizes are 4-12, and dash size is 12-36. These datasets are called \emph{Morse 4.}$\sigma$, $\sigma$ being as before.
    \item Increasing the number of training examples, i.e. the \textbf{size of the dataset}, makes it easier to classify since a NN has more labeled training examples to learn from. Accordingly we chose \emph{Morse 3.1} and scaled the number of examples to obtain \emph{Morse Size} $x$, $x \in \{\sfrac{1}{8},\sfrac{1}{4},\sfrac{1}{2},2,4,8\}$. For example, \emph{Morse Size} $\sfrac{1}{2}$ has 3,500 examples for each class, for a total of 224,000 examples.
\end{enumerate}

\section{Neural Network Results and Analysis}\label{anares}
\subsection{Network Setup}\label{netsetup}
Our NN needs to have 64 output neurons to match the number of classes. The number of input neurons always matches the frame length, i.e. 256 for the \emph{Morse 4.}$\sigma$ datasets, and 64 for all others. We used a single hidden layer with 1024 neurons. The performance, i.e. accuracy, generally increases on adding more hidden neurons, however, we stuck with 1024 since values above that yielded diminishing returns. The network is purely multi-layer perceptron, i.e. there are only fully connected layers. The hidden layer has ReLU activations, while the output is a softmax probability distribution. We used the Adam optimizer with default parameters \cite{Kingma2015}, He normal initialization for the weights \cite{He2015}, and trained for 30 epochs using a minibatch size of 128. We used $\sfrac{6}{7}$th of the total examples for training the NN and the remaining $\sfrac{1}{7}$th for testing at the end of training. All reported accuracies are those obtained on the test samples.

No constraints were imposed on the weights for the NNs training on \emph{Morse 1.}$\sigma$, \emph{2.}$\sigma$ and \emph{3.}$\sigma$, since our experimental results indicated that this led to optimum performance. However, the NNs for \emph{Morse 4.}$\sigma$ are more prone to overfitting due to having more input neurons, leading to more weight parameters. Accordingly we regularized the weights using an L2 coefficient $\lambda = 10^{-5}$, which was the best value as determined experimentally.

\subsection{Results}\label{fcresults}
\begin{figure}[!t]
\centering
\includegraphics[width = 1.0\linewidth]{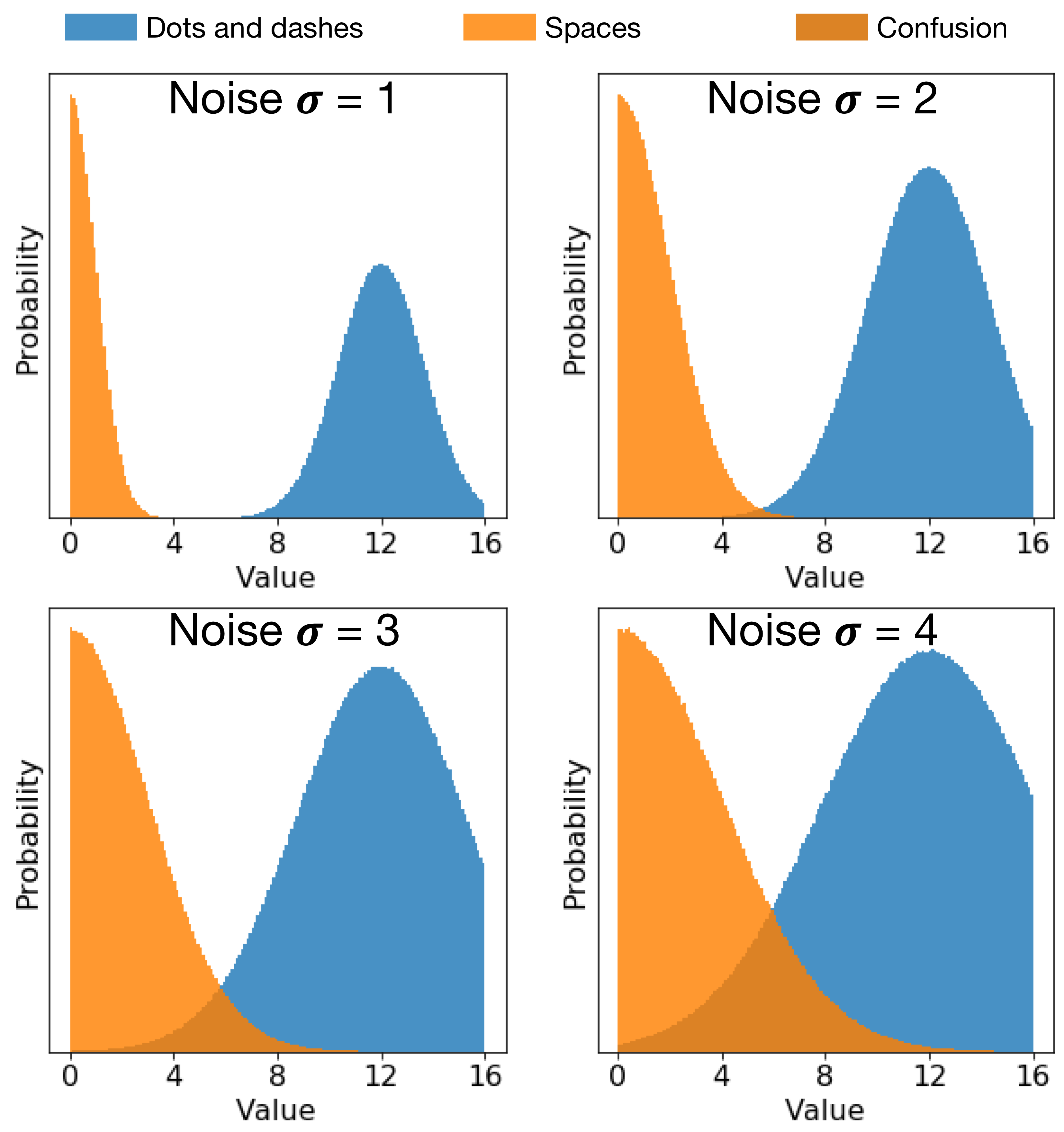}
\caption{Effects of noise leading to spaces (orange) getting confused (brown) with dots and dashes (blue). Higher values of noise $\sigma$ lead to increased probability of the brown region, making it harder for the NN to discern between `dots and dashes' and spaces. The x-axis in each plot shows values in the range $[0,16]$, i.e. before normalizing to $[0,1]$.}
\label{fig-noise}
\end{figure}

Note that the entirety of this work -- generation of various datasets, implementing NNs to process them, and evaluation of metrics -- uses the Python programming language. Test accuracy results after training the NN on the different Morse datasets are shown in Fig. \ref{fig-fc_results}. As expected, increasing the standard deviation of noise results in drop in performance. This effect is not felt strongly when $\sigma=1$ since the $3\sigma$ range can take spaces to a value of 3 (on a scale of $[0,16]$, i.e. before normalizing to $[0,1]$), while dots and dashes can drop to $8-3=5$, so the probability of a space being confused with a dot or dash is basically 0. Confusion can occur for $\sigma\geq2$, and gets worse for higher values, as shown in Fig. \ref{fig-noise}.

Since the codeword lengths do not often stretch beyond 32, the first half of neurons usually encounter high input intensity values corresponding to dots and dashes during training. This means that the latter half of neurons mostly encounter lower input values corresponding to spaces. This aspect changes when introducing leading spaces, which become inputs to some neurons in the first half. The result is an increase in the variance of the input to each neuron. As a result, accuracy drops. The degradation is worse when dashes can have a length of 3-9. Since the lengths are drawn from a uniform distribution, $\sfrac{1}{7}$th of dashes can now be confused with $\sfrac{1}{3}$rd of dots and $\sfrac{1}{3}$rd of intermediate spaces. As an example, for the \textbf{+} codeword which has 2 dashes, 3 dots and 4 intermediate spaces, there is a $\left(\sfrac{2}{9}\times\sfrac{1}{7}+\sfrac{3}{9}\times\sfrac{1}{3}+\sfrac{4}{9}\times\sfrac{1}{3}\right) = 29\%$ chance of this confusion occurring. Dilating by 4, however, reduces this chance to $\left(\sfrac{2}{9}\times\sfrac{1}{25}+\sfrac{3}{9}\times\sfrac{1}{9}+\sfrac{4}{9}\times\sfrac{1}{9}\right) = 9.5\%$. Accuracy is better as a result, and is further improved by properly regularizing the NN so that it doesn't overfit.

\begin{figure}[!t]
\centering
\includegraphics[width = 0.9\linewidth]{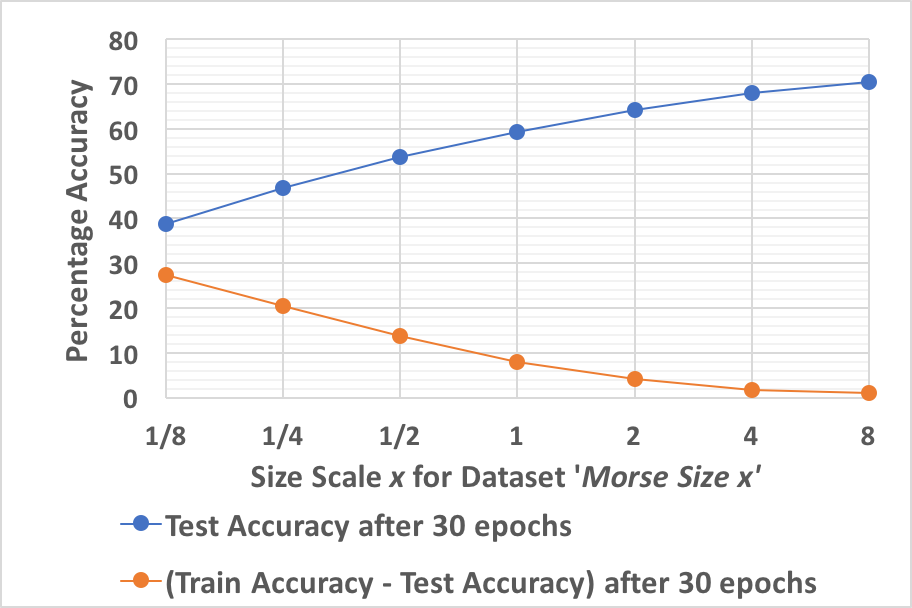}
\caption{Effects of increasing the size of \emph{Morse 3.1} by a factor of \emph{x} on test accuracy after 30 epochs (blue), and (Training Accuracy - Test Accuracy) after 30 epochs (orange).}
\label{fig-dataset_size}
\end{figure}

Increasing dataset size has a beneficial effect on performance. Giving the NN more examples to train from is akin to training on a smaller dataset for more epochs, with the important added advantage that overfitting is reduced. This is shown in Fig. \ref{fig-dataset_size}, which shows improving test accuracy as the dataset is made larger. At the same time, the difference between final training accuracy and test accuracy reduces, which implies that the network is generalizing better and not overfitting. Note that \emph{Morse Size 8} has 3 million labeled training examples -- a beneficial consequence of being able to cheaply generate large quantities of synthetic data.

\subsection{Results for Sparse Networks}\label{sparseresults}
Our previous work \cite{Dey2017_ICANN,Dey2017_Asilomar,Dey2018_ITA} has focused on network complexity reduction in the form of pre-defined sparsity. In a pre-defined sparse network, as opposed to a fully connected one, a fraction of the weights are chosen to be deleted before starting training. These weights never appear during the workflow of the NN. Consider our (64,1024,64) NN as an example. When fully connected, it has $64\times1024+1024\times64=131,072$ weights, which gives a fractional density = 1. If we choose to delete 75\% of the weights at the beginning, then we are left with a NN which has 32,768 weights, i.e. fractional density = $\sfrac{1}{4}$. This leads to reduced storage and operational complexity, which is particularly important for hardware realizations of NNs, but possibly at the cost of performance degradation.

\begin{figure}[!t]
\centering
\includegraphics[width = 1.0\linewidth]{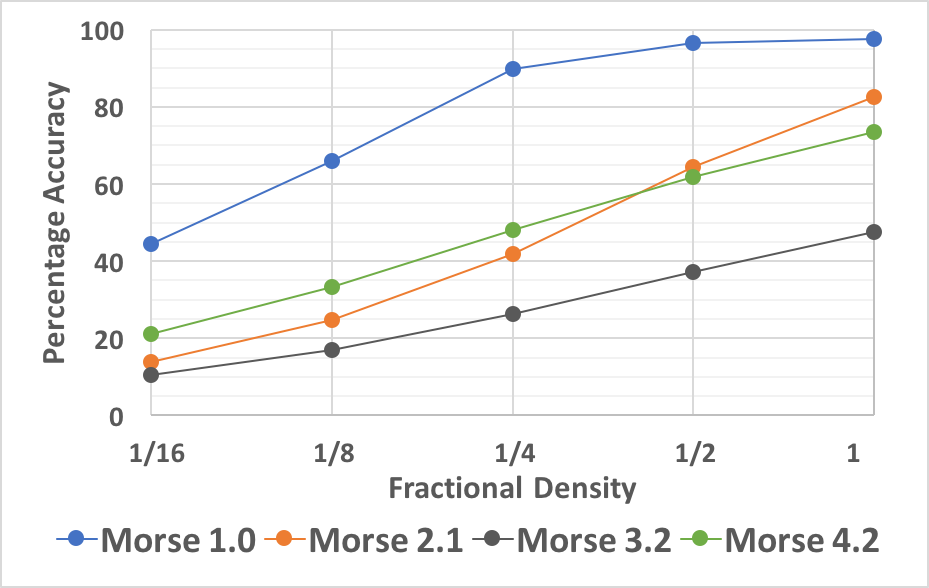}
\caption{Effects of imposing pre-defined sparsity by reducing the density of NNs on performance, for different Morse datasets.}
\label{fig-sparsity}
\end{figure}

Fig. \ref{fig-sparsity} shows the performance degradation for 4 different Morse datasets. Note how the baseline dataset is reasonably accurately classified by a NN with only a quarter of the weights, while performance drops off much more rapidly when dataset variations are introduced. These variations lead to increased information content per neuron per training example. As a result, the reduction in information learning capability as a result of deleting weights is much more severe. Also note that as density is reduced, \emph{Morse 4.2} has the best performance out of the non-baseline models tested in Fig. \ref{fig-sparsity}. This is because it has more weights to begin with, due to the increased number of input neurons. Finally, note that regularization was not applied to any of the sparse models since reducing the number of NN parameters reduces the chances of overfitting, so is in itself a form of regularization.

\section{Metrics}\label{metrics}
This section discusses possible metrics for quantifying how difficult a dataset is to classify. Each sample in a dataset is a point in an $N$-dimensional space, $N$ being the number of features. For the Morse datasets (not considering dilation), $N=64$. There are $M$ classes of points, which is also 64 in our case. The classification problem is essentially finding the class of any new point. Any machine learning classifier will attempt to learn and construct decision boundaries between the classes by partitioning the whole space into $M$ regions. The samples of a particular class $m$ are clustered in the $m$th region. Suppose a particular input sample actually belongs to class $m$. The classifier commits an error if it ranks some class $j$, $j \ne m$, higher than $m$ when deciding where that input sample belongs. The probability of this happening is $P_{PW}\left(j|m\right)$, where subscript $PW$ stands for pairwise and indicates that the quantity is specific to classes $j$ and $m$. The overall probability of error $P(E)$ would also depend on the prior probability $P(m)$ of the $m$th class occurring. Considering all classes in the dataset, $P(E)$ is given according to \cite{Chugg2012_IterativeBook} as:
\begin{IEEEeqnarray}{c}\label{eq-error}
\sum _{m=1}^{M} {P(m) \left[ \max_{ \substack{j\in\{1,2,\cdots,M\} \\ j\ne m} } {P_{PW}\left(j|m\right)} \right]} \leq P(E) \IEEEnonumber \\
\leq \sum _{m=1}^{M} {P(m) \sum _{ \substack{j=1 \\ j\ne m} }^{M} {P_{PW}\left(j|m\right)} }
\end{IEEEeqnarray}

The pairwise probabilities can be approximately computed by assuming that the locations of samples of a particular class $m$ are from a Gaussian distribution with mean located at the centroid $c_m$, which is the average of all samples for the class. To simplify the math, we take the average variance across all $N$ dimensions within a class -- this gives the variance $\sigma_m^2$ for class $m$. The distance between 2 classes $m$ and $j$ is the L2-norm between their centroids, i.e. $d(m,j) = {\left\|c_m-c_j\right\|}_{2}$. A particular class will be more prone to errors if it is close to other classes. This can be quantified by looking at $\frac{d_{min}(m)}{\sigma_m}$, where the numerator is given as:
\begin{IEEEeqnarray}{c}\label{eq-dmin}
d_{min}(m) = \min _{ \substack{j\in\{1,2,\cdots,M\} \\ j \ne m} } {d(m,j)}
\end{IEEEeqnarray}
With the Gaussian assumption, eq. (\ref{eq-error}) simplifies to $L \leq P(E) \leq U$, where:
\begin{IEEEeqnarray}{c}\label{eq-LUmetric}
L = \sum _{m=1}^{M} { P(m) Q\left( \sqrt{\frac{{d_{min}(m)}^2}{4{\sigma_m}^2}} \right) } \IEEEyesnumber \IEEEyessubnumber \\
U = \sum _{m=1}^{M} {P(m) \sum _{ \substack{j=1 \\ j\ne m} }^{M} { Q\left( \sqrt{\frac{d(m,j)^2}{4{\sigma_m}^2}} \right) } } \IEEEnonumber \IEEEyessubnumber
\end{IEEEeqnarray}
where $Q(.)$ is the tail function for a standard Gaussian distribution.

$L$ and $U$ can thus be used as metrics for dataset difficulty, higher values for them imply higher probabilities of error, i.e. lower accuracy. A simpler metric can be obtained by just considering $\frac{\sigma_m}{d_{min}(m)}$. Higher values for this indicate that a) class $m$ is close to some other class and the NN will have a hard time differentiating between them, and b) Variance of class $m$ is high, so it's harder to form a decision boundary to separate inputs having labels $m$ from those with other labels. Since $\frac{\sigma_m}{d_{min}(m)}$ is different for every class, we experimented with ways to reduce it to a single measure such as taking the minimum, the average and the median. The average worked best, which gives our 3rd metric $D$:
\begin{IEEEeqnarray}{c}\label{eq-Dmetric}
D = \frac{ \sum _{m=1}^{M} {\frac{\sigma_m}{d_{min}(m)}} }{M}
\end{IEEEeqnarray}
Therefore high values of $D$ lead to low accuracy.

The 4th and final metric is $T$, to obtain which, we first compute the class centroids just as before. Then we compute the L1-norm between every pair of centroids and average over $N$, i.e:
\begin{IEEEeqnarray}{c}\label{eq-d1}
d_1(m,j) = \frac{{\left\|c_m-c_j\right\|}_{1}}{N}
\end{IEEEeqnarray}
Since all $N$ features in each input sample are normalized to $[0,1]$, all the elements in all the centroid vectors also lie in the range $[0,1]$. So the $d_1$ number for every pair of classes is always between 0 and 1, in fact, it is proportional to the absolute distance between the 2 classes. Then we simply count how many of the $d_1$ numbers are less than a threshold, which we empirically set to 0.05. This gives $T$, i.e.:
\begin{IEEEeqnarray}{c}\label{eq-Tmetric}
T = \sum _{m=1}^{M} { \sum _{\substack{j=1 \\ j\ne m}}^{M} { \mathbb{I} \left( d_1(m,j)<0.05 \right) } }
\end{IEEEeqnarray}
where $\mathbb{I}$ is the indicator function, which is 1 if the condition in its argument is true, otherwise 0. The higher the value of $T$, the lower the accuracy. Note that the total number of $d_1$ values will be $\binom{M}{2}$, so the count for $T$ will typically be higher for datasets that have more classes. This is a desired property since more number of classes usually makes a dataset harder to classify. Note that the maximum value of $T$ for the Morse datasets is $\binom{64}{2}=2016$.

\subsection{Goodness of the Metrics}\label{metric-goodness}
\begin{figure}[!t]
\centering
\includegraphics[width = 1.0\linewidth]{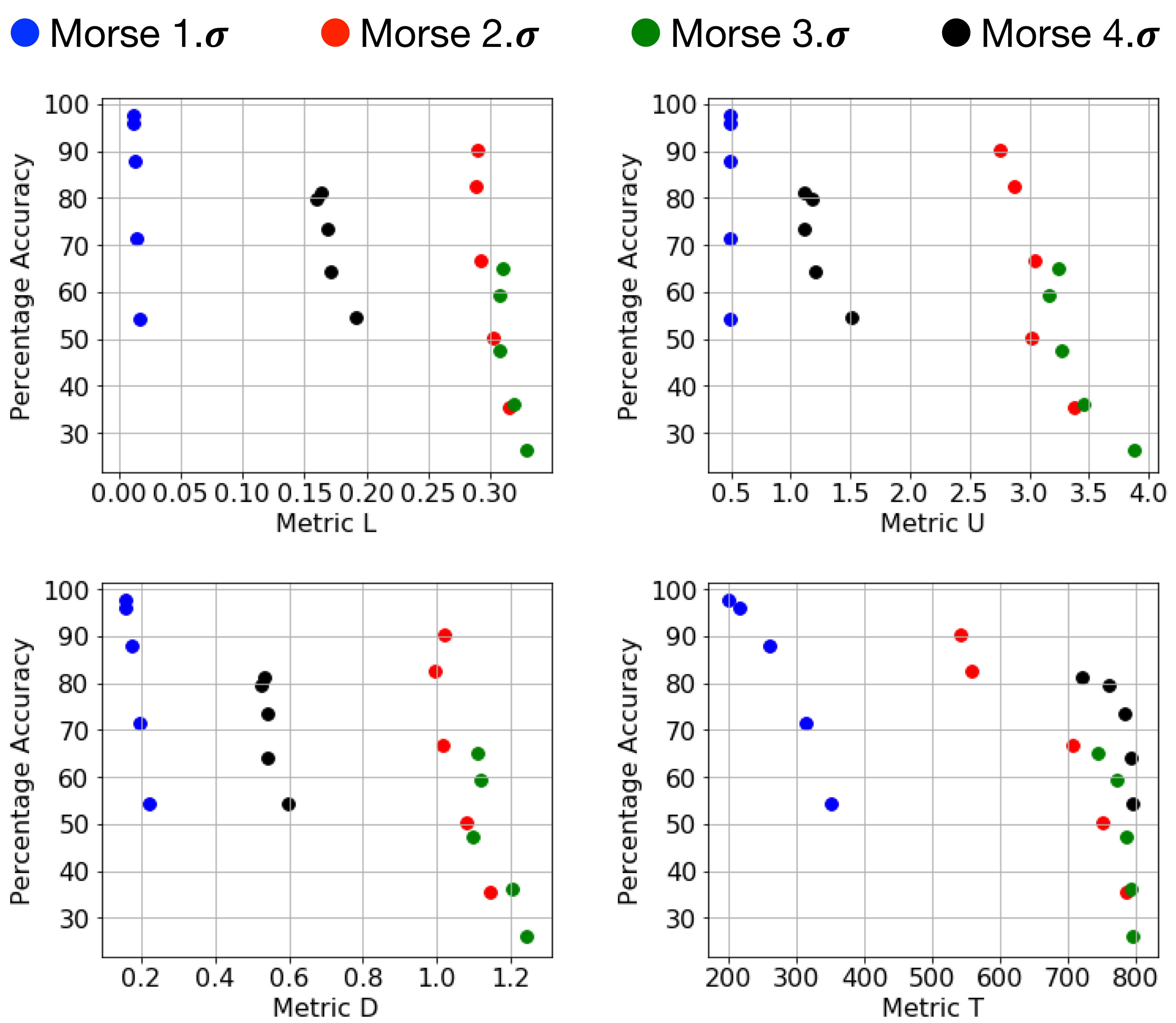}
\caption{Plotting each metric vs. percentage accuracy obtained for datasets \emph{Morse 1.}$\sigma$ (blue), $2.\sigma$ (red), $3.\sigma$ (green) and $4.\sigma$ (black). The accuracy results are using the fully connected network, as reported in Section \ref{fcresults}. Color coding is just for clarity, the $\rho$ values in Table \ref{table-metrics} take into account all the points regardless of color.}
\label{fig-metrics}
\end{figure}

We computed $L$, $U$, $D$ and $T$ values for all the Morse datasets and plotted these with the classification accuracy results obtained from Section \ref{fcresults}. The results are shown in Fig. \ref{fig-metrics}, while the correlation coefficient $\rho$ of each metric with the accuracy is given in Table \ref{table-metrics}. Note that the metrics are an indicator of dataset difficulty, so they are negatively correlated with accuracy. It is apparent that the $U$ and $T$ metrics are the best since their $\rho$ values have the highest magnitude.

\begin{table}[!t]
\renewcommand{\arraystretch}{1.3}
\caption{Correlation coefficients between metrics and accuracy}
\label{table-metrics}
\centering
\begin{tabular}{|c|c|}
\hline
Metric & $\rho$\\
\hline
\hline
$L$ & -0.59\\
\hline
$U$ & -0.64\\
\hline
$D$ & -0.63\\
\hline
$T$ & -0.64\\
\hline
\end{tabular}
\end{table}

\subsection{Limitations of the Metrics}\label{metric-limitations}
As mentioned, each class has a single variance value which is the average variance across dimensions. This is a reasonable simplification to make because
our experiments indicate that the variance of the variance values for different dimensions is small. However, this simplification possibly leads to the error bounds $L$ and $U$ not being sufficiently tight. A possible improvement, involving significantly more computation, would be to compute the $N\times N$ covariance matrix $K_m$ for each class.

It is worthwhile noting that all these metrics are a function of the dataset only and are independent of the machine learning algorithm or training setup used. On the other hand, percentage accuracy depends on the learning algorithm and training conditions. As shown in Fig. \ref{fig-dataset_size}, increasing dataset size leads to accuracy improvement, i.e. the dataset becoming easier, since the NN has more training examples to learn from. However, increasing dataset size drives all the metric values towards indicating higher difficulty. This is because the occurrence of more examples in each class increases its standard deviation $\sigma_m$ and also makes samples of a particular class more scattered, leading to reduced values for $d$ and $d_1$. We hypothesize that these shortcomings of the metrics are due to the fact that most variations of the Morse datasets have a low SNR, while the metrics (the error bounds in particular) are designed for high SNR problems.

\section{Conclusion}\label{conc}
This paper presents an algorithm to generate datasets of varying difficulty on classifying Morse code symbols. While the results have been shown for neural networks, any machine learning algorithm can be tried and the challenge arising from more difficult datasets used to fine tune it. The datasets are synthetic and consequently may not completely represent reality unless statistically verified with real-world tests. However, the different aspects of the generating algorithm help to mimic real-world scenarios which can suffer from noise or other inconsistencies. This work highlights one of the biggest advantages of synthetic data -- the ability to easily produce large amounts of it and thereby improve the performance of learning algorithms. The given Morse datasets are also useful for testing the limits of various learning algorithms and identifying when they fail or possibly overfit/underfit. 

The metrics discussed, while not perfect, can be used to understand the inherent difficulty of the classification problem on any dataset before applying learning algorithms to it. Future work will involve improving the metrics to achieve higher magnitudes of correlation with accuracy, and extension to other types of neural networks and algorithms.



\bibliographystyle{IEEEtran}
\bibliography{IEEEabrv,aaa_main}

\end{document}